\documentclass{article} 
\usepackage{iclr2021_conference,times}

\usepackage{amsmath,amsfonts,bm}

\def\eqref#1{equation~\ref{#1}}

\def\1{\bm{1}}

\DeclareMathAlphabet{\mathsfit}{\encodingdefault}{\sfdefault}{m}{sl}
\SetMathAlphabet{\mathsfit}{bold}{\encodingdefault}{\sfdefault}{bx}{n}

\usepackage{hyperref}
\usepackage{url}
\title{Knowledgebra: An Algebraic Learning Framework for Knowledge Graph}

\author{Tong Yang $^{1,3,\dagger}$, Yifei Wang $^{2,\dagger}$, Long Sha $^{2,\dagger}$, Jan Engelbrecht $^{1}$ and Pengyu Hong $^{2}$ \\\\
$^{1}$ $\quad$ Department of Physics, Boston College, Chestnut Hill, U.S.\\
$^{2}$ $\quad$ Department of Computer Science, Brandeis University, Waltham, U.S.\\
$^{3}$ $\quad$ Machine Learning Centre of Excellence, J.P. Morgan Chase, Hong Kong. \\
$\dagger$ $\quad$ These authors contributed equally to this work.  \\
\texttt{} \\
}

\iclrfinalcopy 

\begin{document}
\maketitle
\begin{abstract}
Knowledge graph (KG) representation learning aims to encode entities and relations into dense continuous vector spaces such that knowledge contained in a dataset could be consistently represented. 
Dense embeddings trained from KG datasets benefit a variety of downstream tasks such as KG completion and link prediction.
However, existing KG embedding methods fell short to provide a systematic solution for the global consistency of knowledge representation.
We developed a mathematical language for KG based on an observation of their inherent algebraic structure, which we termed as \emph{Knowledgebra}.
By analyzing five distinct algebraic properties, we proved that the semigroup is the most reasonable algebraic structure for the relation embedding of a general knowledge graph.
We implemented an instantiation model, \emph{SemE}, using simple matrix semigroups, which exhibits state-of-the-art performance on standard datasets.
Moreover, we proposed a regularization-based method to integrate chain-like logic rules derived from human knowledge into embedding training, which further demonstrates the power of the developed language.
As far as we know, by applying abstract algebra in statistical learning, this work develops the first formal language for general knowledge graphs, and also sheds light on the problem of neural-symbolic integration from an algebraic perspective.
\end{abstract}
\section{Introduction}
Knowledge graphs (KGs) has raised enormous attention among the general artificial intelligence community, which represent human knowledge as a triplet data structure (\textit{head entity}, \textit{relation}, \textit{tail entity}) and can be applied in various downstream scenarios, such as recommendation system \cite{guo2020survey}, question answering \cite{bordes2014open, bordes2014, huang2019knowledge} and information extraction \cite{hoffmann2011knowledge, daiber2013improving}.
It is therefore important to design appropriate knowledge graph embeddings (KGEs) to capture knowledge in the whole dataset with uniform consistency.
A KG consists of two sets: an entity set $\mathcal{E}=\{e_i\}_{i=1}^{N_e}$ and a relation set $\mathcal{R}=\{r_j\}_{j=1}^{N_r}$.
Knowledge is represented as atomic triplets in the following form:
\begin{align}
    (e_i, r, e_j),
\end{align}
which can be interpreted as following: the entity $e_i$ is of the relation $e$ to the entity $e_j$.
For example, the triplet $(\text{\textbf{Washington, D.C.}},\text{\textbf{isCapitalOf}}, \text{\textbf{USA}})$ states that Washington, D.C. is the capital city of the United States.

KGE aims to encode entities and relations of triplet $(e_i, r, e_j)$ into a continuous vector space, i.e., $(\mathbf{e}_i, \mathbf{r}, \mathbf{e}_j)$, associated with an operation $O_{\textbf{R}}(\cdot)$ that ideally maps $\textbf{e}_i$ to $\textbf{e}_j$. 
\footnote{In the current work, we use regular letters to represent the semantic context of entities and relations, and bold letters to represent the high-dimensional array embedding of them.}
Quantitatively, the performance of an embedding model could be roughly measured by the distance between the mapping result, $O[\mathbf{e}_i, \mathbf{r}]$, and the tail entity embedding, $\mathbf{e}_j$, which is based on a given metric, $\mathcal{D}_E$, defined in the entity embedding space.
The representation design of a single triplet is trivial, while the challenge of KGE lies in the fact that different triplets share entities and relations, which requires a uniform representation of all elements that could consistently represent all triplets in a dataset.
With \textbf{TransE} model \cite{transe} as the starting baseline, researchers have explored the problem of KGE in majorly three directions: by implementing different metrics $\mathcal{D}_E$ which controls the efficiency of the entity embedding \cite{fan2014transition, manifoldE, feng2016knowledge, xiao2015transa}; by designing various operations $O[\cdot,\cdot]$ which determines the consistency of the relation embedding \cite{wang2014knowledge, lin2015learning, ji2015knowledge}; by combining KG embedding and logical rules using rule-based regularization or probabilistic model approximation \cite{guo2016jointly, guo2018knowledge, cheng2021uniker, qu2019probabilistic, harsha2020probabilistic, zhang2019can}. All of the directions have rich mathematical structures.

However, a formal analysis from the perspective of mathematics has been lacking for the general KGE tasks, which leaves the modeling design ungrounded.

In this work, we target the second direction, i.e. consistent relation embedding, and develop a formal language for a general KGE problem. 
Specifically, we observe that the consistency issue in relation embeddings directly leads to an algebraic description, and therefore offers an abstract algebra framework for KGE, which we termed as \emph{\textbf{Knowledgebra}}.
The explicit structure of the Knowledgebra is determined by the details of a specific task/dataset, which could differ in five properties: \textit{totality, associativity, identity, invertibility, and commutativity}.
We notice that relations in a general KG should be embedded in a semigroup structure, and hence propose a new embedding model, \textbf{SemE}, which embeds relations as semigroup elements and entities as points in the group action space.
Furthermore, within the language of Knowledgebra, human knowledge about relations\footnote{This is also called logic rules in the following part.} could be expressed by relation compositions.
We, therefore, propose a simple method to directly integrate logic rules of relations with fact triplets to obtain better embedding models with improved performance.
This method also provides a data-efficient solution for tasks with fewer training fact triplets but rich domain knowledge.

Our work is partially motivated by \textbf{NagE}~\cite{nage}, but differ from it significantly in the following aspects.
Firstly,  we deliver a categorical language for KGE problems, which is much more general than NagE with fewer assumptions;
secondly, we  prove that a group structure would be inappropriate for a large class of problems, where the invertibility could not be enforced;
thirdly, beyond a conventional KGE perspective, we adopt a machine reasoning perspective by considering the impact of chain-like logic rules which is traditionally studied in symbolic AI, and therefore shed light on a potential pathway for neural-symbolic integration.

The rest part of the paper is organized in the following ways: 
Sec.~\ref{knowledgebra} introduces the emergent algebra in KG, i.e. Knowledgebra, and proves that a semigroup structure is suited for a general KGE task;
Sec.~\ref{seme} proposes a model, \textbf{SemE}, for general KGE problems, as an instantiation of Knowledgebra, and demonstrates its performance advantage on benchmark datasets;
in Sec.~\ref{logicrules}, we propose a regularization based method to integrate chain-like logic rules into embedding model training, and delivers a case study using a toy dataset where logic rules are easy to be specified;
in the end, we provide a further investigation on the implementation of \textbf{SemE} and discuss potential directions in the future in Sec.~\ref{discussion}, which could exploit more power of the developed algebraic language, Knowledgebra, in knowledge graph applications.

\section{Knowledgebra: An Emergent Algebra in Knowledge Graph}\label{knowledgebra}
In this section, we would analyze a general KG, and demonstrate the emergence of an algebraic structure, which we term as \emph{Knowledgebra}.
The study of algebraic properties in Knowledgebra would produce constraints on KGE modeling.

\subsection{A Categorical Language for Knowledge Graph}\label{sec:category}

As introduced at the beginning, KGs are composed of two sets: $\mathcal{E}$ and $\mathcal{R}$, with entities in $\mathcal{E}$ linked by arrows representing relations in $\mathcal{R}$.
Although knowledge triplets $\{(e_i, r, e_j)\}$ are the elementary atomic components of a KG, the complexity of the KG structure is not present on the triplet level.
It is the set of logic rules that dictates the global consistency of a KG.
Logic rules are the central topic of machine reasoning.
In the machine reasoning field, relations are a special type of predicates, labeled as $\alpha$, with arity 2. 
A logic rule can be expressed as following:
\begin{align}
    \alpha_0 \leftarrow (\alpha_1, \alpha_2, \cdots, \alpha_m),
\end{align}
where each $\alpha_i$ is a predicate with entity variables as arguments.
The above expression means that the head predicate $\alpha_0$ would be iff all body predicates $\{\alpha_i\}_{i=1}^m$ hold.
There is a special type of logic rules, chain-like rules, which has the following form:
\begin{align}
    r_0[e_1, e_{m+1}] \leftarrow (r_1[e_1, e_2], r_2[e_2, e_3], \cdots, r_m[e_m, e_{m+1}]),
\end{align}
where all predicates are of arity 2, and the head argument of the next predicate is always the tail argument of the previous one.
The "cancellation" of intermediate terms $\{e_i\}_{i=2}^m$ implies a compositional definition of the corresponding type of logic rules, where a \emph{composition} of two predicates $r_a$ and $r_b$ is denoted as $r_a\circ r_b$.
Furthermore, it has been proved in \cite{nage} that the composition defined above is associative.

All elements discussed above have indicated the existence of an abstract mathematical structure: \emph{category}.
In mathematics, a category $C$ consists of~\cite{nlab:toposes}:
\begin{itemize}
    \item a class $ob(C)$ of objects;
    \item a class $hom(C)$ of morphisms, or arrows, or maps between the objects;
    \item a domain, or source object class function $dom: hom(C)\rightarrow ob(C)$;
    \item a codomain, or target object class function $cod: hom(C)\rightarrow ob(C)$;
    \item for every three objects a, b and c, a binary operation $hom(a, b)\times hom(b, c)\rightarrow hom(a, c)$ called composition of morphisms; the composition of $f:a\rightarrow b$, and $g: b\rightarrow c$ is written as $g\circ f$;
\end{itemize}
such that the following axioms hold:
\begin{enumerate}
    \item \textbf{Associativity:} if $f:a\rightarrow b$, $g: b\rightarrow c$ and $h: c\rightarrow d$, then $h\circ (g\circ f) = (h\circ g)\circ f$;
    \item \textbf{Identity:} for every object $x$, there exists a morphism $1_x: x\rightarrow x$, called the identity morphism for $x$, such that every morphism $f:a\rightarrow x$ satisfies $1_x\circ f = f$, and every morphism $g : x\rightarrow b$ satisfies $g\circ 1_x = g$.
\end{enumerate}
It is straightforward to examine that all above definitions and axioms hold for a general knowledge graph, which therefore suggests that knowledge graphs naturally host a categorical language description.
In this work, all our later discussions would then utilize concepts and properties of categories, which provide a formal basis.

\subsection{Logic Construction versus Logic Extraction}\label{sec:logic}

With regards to logic rules, there are two pathways in the research of knowledge graphs: namely, \emph{logic construction} and \emph{logic extraction}.

Logic construction is widely used in machine reasoning via symbolic programming, where predicates are built as modules and logic rules are constructed explicitly.
This is similar to the case of a theorem prover, where the propagation from sub-queries to a query is governed by pre-defined rules composed of logical operators, e.g. conjunctions and disjunctions.
Logic construction is a completely deductive process.
With explicit logic construction, one could derive both conclusions and reasoning paths at the same time.
There are several advantages of applying logic construction:
firstly, one could integrate common sense knowledge and domain expertise into the modeling of the reasoning process, which requires less or nearly zero data dependence;
secondly, as rules are constructed explicitly, rigorousness could be guaranteed;
thirdly, with the potential to construct a complete reasoning path, the interpretability of a conclusion derivation could be easily achieved.
On the other side, the disadvantages of logic construction-based approaches are also significant:
\begin{enumerate}
    \item the hand-crafting effort of integrating logic rules becomes impractical when the number of rules gets large;
    \item the explicit construction could not accommodate any possible faults;
    \item the construction could only take into account rules known \emph{a priori}, and could not observe new ones\footnote{With logic operators, higher-order rules could be composed; However, here refer to an inductive process to obtain new elementary rules}.
\end{enumerate}
These problems have been addressed in an alternative method: logic extraction.

Different from logic construction, logic extraction-based approaches belong to the category of statistical learning.
And, opposite to the spirit of logic construction, logic extraction is an inductive process, which infers logic rules implied by a collection of data samples.
From a perspective of reasoning, a redundant set of facts is provided in the phase of modeling, to induce logic rules either explicitly or implicitly, and the inference phase would take in both facts and rules to derive new facts.
KGE is a typical logic extraction-based method that infers logic rules implicitly: the inference phase would produce only a satisfaction score for each query (in the form of a triplet), rather than a complete reasoning path with the usage of explicit logic rules.
Logic extraction could take advantage of huge datasets and is fault-tolerant based on its statistical nature.
Besides, the induction process is insensitive to the number of logic rules and hence scales well with an increasing number of rules.
It is obvious, though, that logic extraction could not integrate with human knowledge easily, and also suffers from the interpretability issue.

It is also noteworthy to emphasize an extra challenge for logic extraction on the implementation level:
the set of logic rules hidden in a dataset requires a \emph{global consistency} of knowledge representation.
In the context of KGE, relation embeddings are not independent of each other and should accommodate all chain-like logic rules under compositions.

\subsection{Algebraic Constraints in KGE}\label{sec:constraint}

With the discussion above, now we consider the problem of KGE.
KGE belongs to the class of logic extraction which explores a given dataset to infer logic rules.
There are two embeddings of KGE, i.e. entity embeddings and relation embeddings.
The chain-like logic rules, however, are entity-independent and only related to relation embeddings.
As introduced above, relations correspond to morphisms in a category and are not independent of each other due to the existence of hidden logic rules under compositions.

The class $hom(C)$, i.e. the set of relations, forms an algebraic structure, which, in general, is termed as \emph{knowledgebra}.
To specify an algebraic structure, the following five properties are usually discussed:
\begin{itemize}
    \item \textbf{totality:} $\forall r_a, r_b \in hom(C)$, $r_a\circ r_b$ is also in $hom(C)$;
    \item \textbf{associativity:} $\forall r_a, r_b, r_c \in hom(C)$, $r_a\circ (r_b \circ r_c) = (r_a\circ r_b) \circ r_c$;
    \item \textbf{identity:} $\exists e\in hom(C), \forall r \in hom(C), e\circ r = r\circ e = r$;
    \item \textbf{invertibility:} $\forall r \in hom(C), \exists \bar{r}\in hom(C), r\circ \bar{r} = \bar{r}\circ r = e$;
    \item \textbf{commutativity:} $\forall r_a, r_b \in hom(C), r_a\circ r_b = r_b\circ r_a$.
\end{itemize}
Variant algebraic structures could be differed by these five properties, and we list 10 well-studied structures in Tab.\ref{tab:properties}.
\begin{table}[!ht]
    \centering
    \begin{tabular}{|c|c|c|c|c|c|}
        \hline
         & totality & associativity & identity & invertibility & commutativity \\
        \hline
        semigroupoid & - & $\checkmark$ & - & - & - \\
        \hline
        small category & - & $\checkmark$ & $\checkmark$ & - & - \\
        \hline
        groupoid & - & $\checkmark$ & $\checkmark$ & $\checkmark$ & - \\
        \hline
        magma & $\checkmark$ & - & - & - & - \\
        \hline
        unital magma & $\checkmark$ & - & $\checkmark$ & - & - \\
        \hline
        loop & $\checkmark$ & - & $\checkmark$ & $\checkmark$ & - \\
        \hline
        semigroup & $\checkmark$ & $\checkmark$ & - & - & - \\
        \hline
        monoid & $\checkmark$ & $\checkmark$ & $\checkmark$ & - & - \\
        \hline
        group & $\checkmark$ & $\checkmark$ & $\checkmark$ & $\checkmark$ & - \\
        \hline
        abelian group & $\checkmark$ & $\checkmark$ & $\checkmark$ & $\checkmark$ & $\checkmark$ \\
        \hline
    \end{tabular}
    \caption{Various algebraic structure and their properties~\cite{enwiki:1053436352}}
    \label{tab:properties}
\end{table}

To fully specify the structure of Knowledgebra, we now examine the five properties in the context of KGE.
An analysis in \cite{nage} claimed all first four properties: totality, associativity, identity, and invertibility, should hold for in KGE modeling, and authors thus developed a group-based framework for relation embeddings.
While we agree with most of the analysis in \cite{nage}, we now provide an argument specifically on the invertibility property.
Consider the following logic rule example consisting of two kinship relations:
\begin{align}\label{eq:conflict1}
    r_a = \text{\textbf{isMotherOf}}, &\quad r_b = \text{\textbf{isBrotherOf}}, \nonumber \\
    r_a\circ r_b &= r_a.
\end{align}
Now if a group structure is used for relation embedding, then there always exist an inverse relation $\bar{r}_a$ for $r_a$, then, based on associativity, we would obtain:
\begin{align}
    r_b = (\bar{r}_a\circ r_a)\circ r_b = \bar{r}_a\circ (r_a\circ r_b) = \bar{r}_a\circ r_a = e,
\end{align}
requiring the relation \textbf{isBrotherOf} to be an identity map that always returns the head entity itself---which is incorrect.
Therefore the existence of $r_a$ should be prohibited.
Another less trivial example consists of the following four kinship relations:
\begin{align}\label{eq:conflict2_1}
    r_a = \text{\textbf{isSonOf}}, \quad r_b = \text{\textbf{isMotherOf}}, \quad r_c = \text{\textbf{isFatherOf}},\quad r_d = \text{\textbf{isBrotherOf}},
\end{align}
which could be related by the following two rules abstractly:
\begin{align}\label{eq:conflict2_2}
    r_a\circ r_b &= r_d, \\
    r_a\circ r_c &= r_d.
\end{align}
Again, if a group embedding is implemented, based on invertibility, i.e. $\bar{r}_a$, and associativity, we would obtain:
\begin{align}
    r_b = (\bar{r}_a\circ r_a)\circ r_b = \bar{r}_a\circ (r_a\circ r_b) &= \bar{r}_a\circ r_d, \\
    r_c = (\bar{r}_a\circ r_a)\circ r_c = \bar{r}_a\circ (r_a\circ r_c) &= \bar{r}_a\circ r_d,   
\end{align}
which then demands directly:
\begin{align}
    r_b = r_c,
\end{align}
an obviously incorrect conclusion.
To simultaneously accommodate the two equations in Eq.\ref{eq:conflict2_2}, the element $r_a$ should not be invertible.
This suggests invertibility is not a desired property for relation embedding in KG.
As in \cite{nage}, the existence of identity element is proved based on invertibility, we could also ignore it for now\footnote{The existence of identity is not necessary but indeed compatible without any conflict.}.
Therefore, in the end, only totality and associativity are natural properties of KGE tasks, which, according to Tab.\ref{tab:properties}, indicates that a semigroup-based relation embedding is desired.

\section{A Semigroup based Instantiation of Knowledge Graph Embedding}\label{seme}
In the above section, we delivered a formal analysis of KGE problems and proved that relations in a KG could generally be embedded in a semigroup structure.
In this section, we implement this proposal by constructing an instantiation model, termed as \textbf{SemE}, and demonstrate the power of algebraic-based embedding on several benchmark tasks.

\subsection{Model Design and Analysis}\label{sec:model}
We firstly introduce the proposed model, including embedding space and distance function design.

\subsubsection{Embedding spaces for entities and relations}
\label{sec:model-nonshare}
We choose the simplest semigroup which has a straightforward parametrization: real $k\times k$ matrices, as the embedding space for relations.
It reduces to $GL(k, \mathbb{R})$ with an extra condition: $det\neq 0$, which guarantees the invertibility.
Entities are embedded as real vectors.
Similar to the implementation in \cite{nage}, to prevent the curse of dimensionality while allowing an embedding space large enough to accommodate  knowledge graphs, we apply block-diagonal matrices as relation embeddings:
\begin{align}\label{large_matrix}
    \setlength{\tabcolsep}{4pt}
    \centering
    \left[
    \begin{tabular}{ccc|ccc|ccc|ccc}
     {} & {} & {} & {} & {} & {} & {} & {} & {} & {} & {} & {} \\
     {} & $M_1$ & {} & {} & $0$ & {} & {} & $\ldots$ & {} & {} & $0$ & {} \\
     {} & {} & {} & {} & {} & {} & {} & {} & {} & {} & {} & {} \\
     \hline
     {} & {} & {} & {} & {} & {} & {} & {} & {} & {} & {} & {} \\
     {} & $0$ & {} & {} & $M_2$ & {} & {} & $\ldots$ & {} & {} & $0$ & {} \\
     {} & {} & {} & {} & {} & {} & {} & {} & {} & {} & {} & {} \\
     \hline
     {} & {} & {} & {} & {} & {} & {} & {} & {} & {} & {} & {} \\
     {} & $\vdots$ & {} & {} & $\vdots$ & {} & {} & $\ddots$ & {} & {} & $\vdots$ & {} \\
     {} & {} & {} & {} & {} & {} & {} & {} & {} & {} & {} & {} \\
     \hline
     {} & {} & {} & {} & {} & {} & {} & {} & {} & {} & {} & {} \\
     {} & $0$ & {} & {} & $0$ & {} & {} & $\ldots$ & {} & {} & $M_n$ & {} \\
     {} & {} & {} & {} & {} & {} & {} & {} & {} & {} & {} & {} 
    \end{tabular}
    \right]
    \left[
    \begin{tabular}{c}
           \\
         $v_1$ \\
           \\
         \hline
          \\
         $v_2$ \\
          \\
         \hline
         \\
         \vdots \\
         \\
         \hline
          \\
         $v_n$ \\
         ${}$
    \end{tabular}
    \right],
\end{align}
where each $M_i$ is a real $k\times k$ matrix and each $v_i$ is a vector in $\mathbb{R}^k$, i.e. entities are embedded in $(\mathbb{R}^k)^{\otimes n}$.
We label the $(nk)\times (nk)$ embedding matrix for relation $r$ as $\mathbf{M}_r$, and the $(nk)$-dim embedding vector for entity $e$ as $\mathbf{v}_e$.

\subsubsection{Distance function for similarity measure}
To apply an end-to-end gradient-based training, a distance function to compare the similarity between two arbitrary entities is required.
The two most common choices are Euclidean distance and cosine distance.
The latter one, i.e. cosine similarity, focuses only on the relative angle between two high dimensional vectors while ignoring the radial component.
In the current work, a general $k\times k$ matrix transform a $k$-dim vector in 6 ways, including 5 affine-type transformations: translations, rotations, reflections, scaling maps, and shear maps, and projections achieved by non-invertible matrices, most of which, except rotations and reflections, cannot be differed by the cosine similarity, and we, therefore, choose Euclidean distance to measure entity similarity.
For a fact triplet $(e_i, r, e_j)$, the performance of a SemE model would therefore be measured by the following similarity measure:
\begin{align}
    s_r(e_i, e_j) = \|\mathbf{M}_r\mathbf{v}_{e_i} - \mathbf{v}_{e_j}\|_2,
\end{align}
where $\|\cdot\|_2$ calculates the $L_2$-norm of a vector.

The complete loss function is designed as follows:
\begin{align}\label{L0}
    &\mathcal{L}_0 =-\frac{1}{p_{\text{loss}}+1}(\log{\sigma\left[\gamma-s_{r}(e_i, e_j)\right]}\nonumber\\
    &\qquad +p_{\text{loss}}\sum_{k=1}^{n} p(e_{ik}^{\prime}, r, e_{jk}^{\prime}) \log{\sigma[s_r(e_{ik}^{\prime}, e_{jk}^{\prime})-\gamma]} \nonumber)\\
    &p\left(e_{ik}^{\prime}, r, e_{jk}^{\prime} |\left\{e_i, r, e_j\right\}\right) =\frac{e^{\alpha[\gamma-s_r(e_{ik}^{\prime}, e_{jk}^{\prime})]}}{\sum_{l} e^{\alpha [\gamma-s_r(e_{il}^{\prime}, e_{jl}^{\prime})]}}
\end{align}
where $\sigma$ is the Sigmoid function, and $\gamma$ is a hyper-parameter controlling the margin to prevent over-fitting, $p_{\text{loss}}$ is a hyper-parameter controlling the ratio of negative and positive losses.
We apply a popularly implemented~\cite{nage, rotate} negative sampling setup, termed as \emph{self-adversarial negative sampling}~\cite{rotate}, with
$e_{ik}^{\prime}$ and $e_{jk}^{\prime}$ being the negative samples for head and tail entity, respectively, while $p(e_{ik}^{\prime}, r, e_{jk}^{\prime})$ is the adversarial sampling weight with the inverse temperature $\alpha$ controlling the focus on poorly learnt samples.

\subsubsection{Low dimensional relation embedding}
\label{sec:shared-block}
In the standard SemE, relations are embedded as $n$ blocks of $k\times k$ matrices while entities are mapped to $(nk)$-dim vectors.
In practice, there are tasks where only simple relations are involved.
For example, the WordNet-18 dataset includes only 18 distinct relations, connected by very simple logic rules.
However, the large number of entities requires a relatively high dimensional vector space for embedding, which easily results in large redundancy in relation embedding in such tasks.
To improve the efficiency of parametrization for tasks with simple relations, and accelerate learning convergence at the same time, we propose two simplified alternatives for relation embedding:
\begin{itemize}
    \item \textbf{shared blocks:} instead of using $n$ distinct $k\times k$ matrices, we use identical copies of one $k\times k$ matrix $M_0$:
    \begin{align}\label{shared_block}
    \setlength{\tabcolsep}{4pt}
    \centering
    \left[
    \begin{tabular}{ccc|ccc|ccc|ccc}
     {} & {} & {} & {} & {} & {} & {} & {} & {} & {} & {} & {} \\
     {} & $M_0$ & {} & {} & $0$ & {} & {} & $\ldots$ & {} & {} & $0$ & {} \\
     {} & {} & {} & {} & {} & {} & {} & {} & {} & {} & {} & {} \\
     \hline
     {} & {} & {} & {} & {} & {} & {} & {} & {} & {} & {} & {} \\
     {} & $0$ & {} & {} & $M_0$ & {} & {} & $\ldots$ & {} & {} & $0$ & {} \\
     {} & {} & {} & {} & {} & {} & {} & {} & {} & {} & {} & {} \\
     \hline
     {} & {} & {} & {} & {} & {} & {} & {} & {} & {} & {} & {} \\
     {} & $\vdots$ & {} & {} & $\vdots$ & {} & {} & $\ddots$ & {} & {} & $\vdots$ & {} \\
     {} & {} & {} & {} & {} & {} & {} & {} & {} & {} & {} & {} \\
     \hline
     {} & {} & {} & {} & {} & {} & {} & {} & {} & {} & {} & {} \\
     {} & $0$ & {} & {} & $0$ & {} & {} & $\ldots$ & {} & {} & $M_0$ & {} \\
     {} & {} & {} & {} & {} & {} & {} & {} & {} & {} & {} & {} 
    \end{tabular}
    \right]
    \left[
    \begin{tabular}{c}
          \\
         $v_1$ \\
          \\
         \hline
          \\
         $v_2$ \\
          \\
         \hline
         \\
         \vdots \\
         \\
         \hline
          \\
         $v_n$ \\
         ${}$
    \end{tabular}
    \right].
    \end{align}
    The number of parameters of embedding for one relation then reduces from $n\times k\times k$ to $k\times k$, which is a super low dimensional embedding, termed as \textbf{SemE-s}.
    \item \textbf{shared blocks with shift:} in the case where a single $k\times k$ matrix is insufficient while low-dimensional efficiency is still demanded, we could break the symmetry among $n$ subspaces by introducing a block-dependent shift $\delta_i$:
    \begin{align}\label{shared_block_shift}
    \setlength{\tabcolsep}{4pt}
    \centering
    \left[
    \begin{tabular}{ccc|ccc|ccc|ccc}
     {} & {} & {} & {} & {} & {} & {} & {} & {} & {} & {} & {} \\
     {} & $M_0$ & {} & {} & $0$ & {} & {} & $\ldots$ & {} & {} & $0$ & {} \\
     {} & {} & {} & {} & {} & {} & {} & {} & {} & {} & {} & {} \\
     \hline
     {} & {} & {} & {} & {} & {} & {} & {} & {} & {} & {} & {} \\
     {} & $0$ & {} & {} & $M_0$ & {} & {} & $\ldots$ & {} & {} & $0$ & {} \\
     {} & {} & {} & {} & {} & {} & {} & {} & {} & {} & {} & {} \\
     \hline
     {} & {} & {} & {} & {} & {} & {} & {} & {} & {} & {} & {} \\
     {} & $\vdots$ & {} & {} & $\vdots$ & {} & {} & $\ddots$ & {} & {} & $\vdots$ & {} \\
     {} & {} & {} & {} & {} & {} & {} & {} & {} & {} & {} & {} \\
     \hline
     {} & {} & {} & {} & {} & {} & {} & {} & {} & {} & {} & {} \\
     {} & $0$ & {} & {} & $0$ & {} & {} & $\ldots$ & {} & {} & $M_0$ & {} \\
     {} & {} & {} & {} & {} & {} & {} & {} & {} & {} & {} & {} 
    \end{tabular}
    \right]
    \left[
    \begin{tabular}{c}
          \\
         $v_1$ \\
          \\
         \hline
          \\
         $v_2$ \\
          \\
         \hline
         \\
         \vdots \\
         \\
         \hline
          \\
         $v_n$ \\
         ${}$
    \end{tabular}
    \right]+
    \left[
    \begin{tabular}{c}
          \\
         $\delta_1$ \\
          \\
         \hline
          \\
         $\delta_2$ \\
          \\
         \hline
         \\
         \vdots \\
         \\
         \hline
          \\
         $\delta_n$ \\
         ${}$
    \end{tabular}.
    \right]
    \end{align}
    The number of parameters is then $k\times k + n\times k$. 
    And we term the resulting model as \textbf{SemE-$\delta$s}~\footnote{Importantly, this shift corresponds to a translation in each subspace, which, together with the matrix multiplication, still hold a semigroup structure. The resulting operation is quite similar to a Euclidean group but with non-invertible elements.}.
\end{itemize}
We would implement these low-dim embedding methods later on tasks with simple relations.

\subsection{Experiments on Benchmark Datasets}\label{sec:exp}
\subsubsection{Experimental setup}
\textit{Datasets:} we evaluate the proposed approach on two popular public knowledge graph benchmarks: WN18RR \citep{dettmers2018convolutional} and FB15k-237 \citep{toutanova2015observed}. These two dataset were derived from WN18 \citep{miller1995wordnet} and FB15K \citep{bollacker2008freebase} respectively. The FB15k dataset extracted all FreeBase entities that have over 100 mentions and are featured in the Wikilinks database while the WN18 dataset extracted from a linguistic knowledge graph ontology named the WordNet. After finding the FB15k and the WN18 dataset suffered from test leakage issues due to the presence of equivalent inverse relations, the WN18RR and FB15k-237 were created as more challenging datasets removing all equivalent and inverse relations. In these two datasets, none of the triplets in the training set are directly linked to the validation and test sets.

\textit{Evaluation Metrics}: similar to previous work, we use two ranking-based metrics for evaluation: (1) Cut-off Hit ratio (H@$N$, $N\in\{1, 3, 10\}$), which measures the proportion of correct entity predictions among the top $N$ prediction result cut-off. (2) Mean Reciprocal Rank (MRR), which represents the average of inverse ranks assigned to correct entities. 

\textit{Implementation Details}: we implement our models via the Pytorch framework\footnote{https://pytorch.org/} and experimented on a server with an NVIDIA Tesla V100 GPU (32 GB). 
The Adam optimizer \citep{kingma2014adam} is used with default settings of $\beta_{1}$ and $\beta_{2}$. 
We use a learning rate annealing schedule that discounted the learning rate by a factor of 0.1 with a patience setting of 10. 
The batch size is fixed at 1000.

\subsubsection{Experiment results}

For FB15k-237, we implement the standard \textbf{SemE} as stated in Section \ref{sec:model-nonshare}, with parameterization of $k=5$ and $n=240$. 
Other hyper-parameters are tuned as following: learning rate $\eta \in \{3e-4, 1e-3\}$;  
number of negative samples during training $n_{\text{neg}}\in \{64, 128\}$; 
adversarial negative sampling temperature $\alpha \in \{0.75, 0.85, 0.95, 1\}$; 
loss function margin $\gamma \in \{9, 12\}$; ratio between negative and positive losses $p_{\text{loss}} \in \{5, 10\}$. 
The best model is under configuration of $\eta=1e-3, n_{\text{neg}}=64, \alpha=0.85, \gamma=9, p_{\text{loss}}=5$.
As another benchmark dataset, WN18RR only includes simple relations that can be sufficiently captured by low-dimensional embeddings. 
Therefore we apply a low-dim alternative model, \textbf{SemE-$\delta$s}, as discussed in Section \ref{sec:shared-block}, where we take $k=10$ and $n=100$ in this case. 
Other hyper-parameters of grid search include: $\eta \in \{3e-4, 1e-3\}$; $n_{\text{neg}}\in \{64, 128\}$; $\alpha \in \{0.5, 0.7, 0.85, 1\}$; $\gamma \in \{6, 7, 7.5\}$; $p_{\text{loss}} \in \{10, 20, 30\}$. 
The best performance appears in the configuration with $\eta=1e-3, n_{\text{neg}}=128, \alpha=0.7, \gamma=6, p_{\text{loss}}=30$.
Experimental results of the best models are exhibited in Tab.\ref{tab:hiexp}.

\begin{table}[!ht]
    \begin{center}
    \begin{tabular}{|c|cccc|cccc|}
    \hline
    Model & \multicolumn{4}{c|}{WN18RR}  & \multicolumn{4}{c|}{FB15k-237} \\ \cline{2-9}
    & MRR & H@1 & H@3 & H@10 & MRR & H@1 & H@3 & H@10  \\
    \hline
    \small{TransE \cite{transe}} & 0.226 & - & - & 0.501 & 0.294 & - & - & 0.465 \\
    \small{ComplEx \cite{trouillon2016complex}} & 0.440 & 0.410 & 0.460 & 0.510 & 0.247 & 0.158 & 0.275 & 0.428 \\ 
    \small{DistMult \cite{dismult}} & 0.430 & 0.390 & 0.440 & 0.490 & 0.241 & 0.155 & 0.263 & 0.419 \\ 
    \small{ConvE \cite{dettmers2018convolutional}} & 0.430 & 0.400 & 0.440 & 0.520 & 0.325 & 0.237 & 0.356 & 0.501 \\
    \small{MuRE~\footnotemark \cite{balavzevic2019tucker}}  & 0.475 & \underline{0.436} & 0.487 & 0.554 & 0.336 & 0.245 & 0.370 & 0.521 \\
    \small{RotatE \cite{rotate}} & 0.476 & 0.428 & 0.492 & \underline{0.571} & 0.338 & 0.241 & 0.375 & \underline{0.533} \\ 
    \small{NagE \cite{nage}} & \underline{0.477} & 0.432 & \underline{0.493} & \textbf{0.574} & \underline{0.340} & \underline{0.244} & \underline{0.378} & 0.530 \\ \hline
    \small{\textbf{SemE}} & \textbf{0.481} & \textbf{0.437} & \textbf{0.499} & 0.567 & \textbf{0.354} & \textbf{0.258} & \textbf{0.393} & \textbf{0.548} \\\hline
    \end{tabular}
    \caption{\label{tab:hiexp} Experiment results on WN18RR and FB15k-237 datasets (best scores are marked as bold while second best are underlined). A standard \textbf{SemE} model is applied for FB15k-237, while a low-dim alternative \textbf{SemE-$\delta$s} is used for WN18RR.}
    \end{center}
\end{table}
\footnotetext{This is the Euclidean analogue of MuRP \cite{balavzevic2019tucker}.}
As shown above, \textbf{SemE} models have achieved state-of-the-art performance in all nearly evaluation metrics on the two standard benchmark dataset.
Remarkably, in the task of WN18RR, the low-dimensional model \textbf{SemE-$\delta$s} has already provided promising results with even a small amount of parameters in relation embedding, which further demonstrates the advantage of the proposed approach.

\section{Integrating Human Knowledge into Knowledge Graph Embedding}\label{logicrules}

\subsection{A Regularization Method for Logic Rules}\label{sec:regularization}

One of the major challenges in KGE is to integrate human knowledge, either commonsense or domain knowledge, into the embedding model.
As introduced in Sec.~\ref{sec:logic}, knowledge is expressed as logic rules, which in turn is represented by relation compositions.
The task of integrating human knowledge is equivalent to enforcing the compositional dependence among embeddings of different relations.
For example, the two relations $r_a=\text{\textbf{isWifeOf}}$ and $r_b=\text{\textbf{isHusbandOf}}$ are mutually depend on each other as:
\begin{align}
    r_a\circ r_b = E,
\end{align}
where $E$ is an identity mapping.
When a matrix embedding is implemented, the following equation should hold:
\begin{align}
    \mathbf{M}_{r_a}\cdot \mathbf{M}_{r_b} = \mathbb{I},
\end{align}
which results into an identity matrix.
The above example inspires a way to integrate human knowledge, i.e. logic rules, into embeddings: 
to design an additional loss term that minimizes the matrix distance suggested by rules.
For the instance above, we may add the following term into loss function:
\begin{align}
    \mathcal{L} = \mathcal{L}_0 + \lambda\|\mathbf{M}_{r_a}\cdot \mathbf{M}_{r_b} - \mathbb{I}\|_2,
\end{align}
where $\mathcal{L}_0$ is the usual training loss defined in Eq.\ref{L0}, while the second term regularizes the embeddings of $r_a$ and $r_b$ to be mutually dependent.
In general, for chain-like rules, which could be captured as compositions, we could apply the following regularized loss function:
\begin{align}
    \mathcal{L} = \mathcal{L}_0 + \sum_{i=1}^K
    \lambda_i\|\mathbf{M}_{i, 1}\cdot\mathbf{M}_{i, 2} - \mathbf{M}_{i, 3}\|_2,
\end{align}
where, without loss of generality as argued in the \emph{inverse and compositional hyper-relation}, logic rules are expressed as a compositional dependency of three relations, with one of which could be an identity to capture the case of the inverse.
This provides an efficient approach to integrating human knowledge, i.e. logic rules, into KG embedding tasks.

\subsection{Kinship: A Case Study of Logic Integration}\label{sec:kinship}

We now demonstrate the above proposed regularized loss method on a toy dataset: Hinton's \emph{Kinship} dataset. 
There are 12 relations in this toy KG: \emph{wife, husband, father, mother, son, daughter, sister, brother, uncle, aunt, niece,} and \emph{nephew}.
From commonsense knowledge, we consider the following set of constraints for relation embedding:
\begin{align}
    \text{\textbf{isWifeOf}}\cdot\text{\textbf{isHusbandOf}} &= \mathbb{I}, \\
    \text{\textbf{isHusbandOf}}\cdot\text{\textbf{isMotherOf}} &= \text{\textbf{isFatherOf}}, \\
    \text{\textbf{isSonOf}}\cdot\text{\textbf{isMotherOf}} &= \text{\textbf{isBrotherOf}}, \\
    \text{\textbf{isSonOf}}\cdot\text{\textbf{isFatherOf}} &= \text{\textbf{isBrotherOf}}, \\
    \text{\textbf{isBrotherOf}}\cdot\text{\textbf{isFatherOf}} &= \text{\textbf{isUncleOf}}, \\
    \text{\textbf{isBrotherOf}}\cdot\text{\textbf{isMotherOf}} &= \text{\textbf{isUncleOf}}, \\
    \text{\textbf{isSisterOf}}\cdot\text{\textbf{isFatherOf}} &= \text{\textbf{isAuntOf}}, \\
    \text{\textbf{isSisterOf}}\cdot\text{\textbf{isMotherOf}} &= \text{\textbf{isAuntOf}} \\
    \text{\textbf{isSonOf}}\cdot\text{\textbf{isBrotherOf}} &= \text{\textbf{isNieceOf}} \\
    \text{\textbf{isSonOf}}\cdot\text{\textbf{isSisterOf}} &= \text{\textbf{isNieceOf}} \\
    \text{\textbf{isDaughterOf}}\cdot\text{\textbf{isBrotherOf}} &= \text{\textbf{isNephewOf}} \\
    \text{\textbf{isDaughterOf}}\cdot\text{\textbf{isSisterOf}} &= \text{\textbf{isNephewOf}}
\end{align}
We implemented the logic regularized loss method using a small shared block model, \textbf{SemE-s}, with 2 copies of 2 dimensional subspaces. We used the batch size of 5 for training and 12 for testing. For other hyper-parameters we took $\alpha=0.1,  n_{\text{neg}}=4, p_{\text{loss}}=2$. We set all regularization parameters $\lambda_i = 0.1,\; \forall i$, and compared with the baseline model with $\lambda_i=0,\;\forall i$.
Experimental results on testing dataset are shown in Tab.\ref{tab:demo}.
\begin{table}[!ht]
    \centering
    \begin{tabular}{|c|c|c|c|c|c|}
        \hline
        Model & MR & MRR & $H@1$ & $H@3$ & $H@10$\\
        \hline
        regular \textbf{SemE-s} & 4.83 & 0.464 & 0.292 & 0.458 & 0.875 \\
        \hline
        regularized \textbf{SemE-s} & \textbf{3.71} & \textbf{0.574} & \textbf{0.458} & \textbf{0.583} & \textbf{0.958} \\
        \hline
    \end{tabular}
    \caption{Testing results on regular \textbf{SemE-s} model and logic regularized \textbf{SemE-s} model.}
    \label{tab:demo}
\end{table}

The performance advantage of the logic regularized model is significant, which demonstrates the power of the regularization brought by logic rules.
This showcases an efficient way to integrate external knowledge.

\section{Discussion}\label{discussion}

The mutual dependence of relations in a knowledge graph suggests the existence of an algebraic structure, which is introduced in this work as \emph{Knowledgebra}.
By analyzing a general KG based on five distinct properties, we determined that the semigroup is the most reasonable algebraic structure for general relation embeddings, where only totality and associativity are required.
Our theoretical analysis was based on the work of NagE~\cite{nage}, and differed from it majorly by demonstrating that invertibility should be allowed to break.
In Sec.~\ref{sec:constraint}, we provided proof based on contradictions with several examples among kinship relations.
With the instantiation model, SemE proposed, we could discuss the invertibility issue from an alternative perspective.

SemE applies matrices to embed relations.
A non-invertible matrix $\mathbf{M}$ has a determinant $det(\mathbf{M}) = 0$, which, from a dimensional-perspective, suggests a projection associated with a dimension reduction.
For the example in Eq.~\ref{eq:conflict2_1}, the relation $r_a = \text{\textbf{IsMotherOf}}$ should not be invertible for a family with multiple children, since all vectors corresponding to a \textit{child} should be simultaneously mapped by the matrix $\mathbf{M}_{r_a}$ to the same vector which represents their mother.
In other words, the non-invertible elements in a semigroup are used to capture $N$-to-$1$ relations, which commonly exist in real-life datasets.

A derived question from the above discussion would be the representation of $1$-to-$N$ relations.
Within the context of KGE using statistical learning-based representation, it is challenging to directly design a mathematically rigorous $1$-to-$N$ mapping operation $O(\cdot,\cdot)$, as it always produces a deterministic result.
However, this could be relieved by noting that the final performance of an embedding model is determined not directly by the mapping output but by the ranking of closeness between the output with each candidate entity.
Therefore, instead of producing multiple results, the distributed learning framework requires the output to be as equidistant to all correct candidates as possible.
This also explains the necessity of high-dimensional entity embedding: within a low-dimensional vector space, it is more challenging to find a point equidistant w.r.t multiple points.

With the proposal logic-regularized-loss method in Sec.~\ref{logicrules}, the proposed algebraic learning framework sheds new light on the area of neural-symbolic integration.
More specifically, we proposed a method to integrate chain-like logic rules of relations into distributed representations.
However, this only covers a small set of general logic, and it is, therefore, interesting to develop further methods to integrate other types of logic rules, including ones concerning entity attributes~\footnote{It is also called arity-1 relations.}.
Furthermore, the current work focuses merely on relation embeddings, which have an algebraic nature.
The entity embedding, on the other hand, plays the role of "action space" of the relational algebra and therefore has a geometric nature.
Given an algebraic structure, the choice of its "action space" is far from being fully determined.
There is hence a rich set of candidates for entity embedding design, which is worth to be investigated in the future.

\bibliography{iclr2021_conference}
\bibliographystyle{iclr2021_conference}

\appendix
\end{document}